\documentclass{Interspeech2024}
\usepackage{xcolor}

\interspeechcameraready

\title{Cross-Lingual Conversational Speech Summarization with Large Language Models}

\name{Max}{Nelson$^*$}
\name{Shannon}{Wotherspoon$^*$}
\name{Francis}{Keith}
\name{William}{Hartmann}
\name{Matthew}{Snover}
\address{BBN Technologies, Cambridge MA, USA}
\email{\{max.nelson, shannon.wotherspoon, francis.keith, william.hartmann, matt.snover\}@rtx.com}


\begin{document}

\maketitle
\def\thefootnote{*}\footnotetext{Equal Contribution}\def\thefootnote{\arabic{footnote}}

\begin{abstract}
Cross-lingual conversational speech summarization is an important problem, but suffers from a dearth of resources.
While transcriptions exist for a number of languages, translated conversational speech is rare and datasets containing summaries are non-existent.
We build upon the existing Fisher and Callhome Spanish-English Speech Translation corpus by supplementing the translations with summaries.
The summaries are generated using GPT-4 from the reference translations and are treated as ground truth.
The task is to generate similar summaries in the presence of transcription and translation errors.
We build a baseline cascade-based system using open-source speech recognition and machine translation models.
We test a range of LLMs for summarization and analyze the impact of transcription and translation errors.
Adapting the Mistral-7B model for this task performs significantly better than off-the-shelf models and matches the performance of GPT-4.
    
\end{abstract}

\section{Introduction}
\label{sec:intro}

Despite the advances in automatic speech recognition (ASR) since the advent of deep neural networks, conversational speech remains a significant challenge.
Due to the lack of data and challenging recording conditions, word error rates (WER) remain high.
Even when accurately transcribed, conversational speech is difficult to read; making it beneficial to build models of cross-lingual summarization that can generate more human-readable versions of conversations. 
Faithful summarization captures the important information in the conversation without the distractions of hesitations and other speech disfluencies.

Recent advances in large language models (LLMs) have allowed them to match or even surpass the capabilities of special purpose models for a number of tasks, including summarization. 
For some datasets, LLMs have been found to not only surpass previous summarization models, but to meet the performance of humans \cite{zhang2024benchmarking}.
LLMs have the additional benefit of flexibility.
With prompting and finetuning, the model can be made to take advantage of additional information, or to provide a contextual summary based on additional instructions.

Summarization of speech has a long history \cite{valenza1999summarisation}.
The majority of the work has focused on single-speaker audio with a clear goal (e.g., voicemail \cite{koumpis2000transcription}, broadcast news \cite{hori2002automatic}).
Two-party conversations have also been a domain of interest.
Meeting summarization was explored in \cite{murray05extractive}.
\cite{sharma2023espnet} released a large corpus of interviews where a host interviews a guest.
Another major area is the summarization of call center interactions \cite{zou2021topic}.
Given the audio from an interaction between a customer and a call center employee, the goal is to describe the nature of the call and whether and how a request was fulfilled.
To our knowledge, there is no prior work on the cross-lingual summarization of conversational speech outside of the call-center domain.

The classic approach to cross-lingual speech summarization has been a cascaded pipeline where audio is automatically transcribed and then fed to a summarization system \cite{furui2004speech}.
A benefit of this approach is that the individual models can be trained independently, taking advantage of non-parallel data.
More recent work has explored direct summarization where a single model is used to directly summarize the audio \cite{kano2023summarize}.
The model can be directly optimized for the task, as opposed to individual components being optimized for intermediate objectives.

Summarization can be either abstractive or extractive.
In a cross-lingual conversational speech domain, extractive summarization can be problematic.
Both transcription and translation errors are propagated, and the extracted utterances can be incomplete or incoherent \cite{liu2013towards}.
Because of these issues, we focus on abstractive summarization.
While direct approaches can be powerful, we focus on a cascaded approach which allows us to incorporate and compare open-source models.
We aim to establish an evaluation framework for conversational speech summarization and to evaluate the ability of LLMs to accomplish the task.
We leave comparisons against direct summarization to future work.
Our contributions are as follows:

\begin{itemize}
    \item We provide a first of its kind public conversational speech summarization dataset\footnote{https://github.com/hartmannw/spanish-cts-summarization} by building upon existing datasets.
    \item We compare a range of LLMs and provide baseline performance using open-source tools and models.
    \item We demonstrate that by fine-tuning a relatively small, quantized LLM, we achieve performance competitive with GPT-4.
\end{itemize}

\section{Technical Approach and Dataset Creation}
\label{sec:approach}

\subsection{Datasets}
\label{sec:data}

We focus on the Fisher and Callhome corpora of Spanish conversational telephone speech (CTS).
The audio and Spanish transcripts of which are available through the LDC\footnote{https://catalog.ldc.upenn.edu/\{LDC2010T04, LDC2010S01, LDC96T17, LDC96S35\}}.
The corpora contain crowd-sourced English translations \cite{post2013improved} of Spanish telephone calls.
The translations are publicly available\footnote{https://github.com/joshua-decoder/fisher-callhome-corpus}.
We focus on this dataset due to the large amount of transcribed audio compared to other CTS datasets and the availability of translations.
We are unaware of similar CTS datasets with English translations.
The existence of English reference translations are critical as they are used for reference summary generation.

The Callhome corpus comes with a predefined train/test split.
For the Fisher data, we adopt the data splits defined by Post et al. \cite{post2013improved} in order to align with their translations and results.
We report results across both Callhome test splits (Devtest, Evltest) and all three Fisher test splits (Dev, Dev2, Test).

\subsection{Summary Generation}
\label{sec:summary}

While we have human generated translations for this dataset, there are no existing summaries.
The collection of human summaries for this dataset would be expensive and time-consuming.
Instead, we generate summaries using GPT-4\footnote{https://openai.com/research/gpt-4}.
We justify this decision in two ways.
The first is that summaries generated by GPT-4's predecessors have been judged comparable to human summaries on some datasets\cite{zhang2024benchmarking}.
Given the difficulty of summarizing conversational speech, there is no guarantee that summaries generated by humans would be substantially better.
Second, GPT-4 is given access to reference translations when generating the summaries.
During evaluation, an LLM will be given input that contains both ASR and machine translation (MT) errors.
The goal is to generate a summary from errorful input that can match the reference summary.
Even if the reference summaries are deficient, obtaining a similar result in the presence of errors would still signify a significant achievement.

\begin{table}
    \caption{\label{tab:dataset} {\it Breakdown of the number of conversations (Conv.), chunks, utterances, and hours of audio (Hrs) across the datasets. Summarization happens at the level of chunks.}}
    \centering
    \begin{tabular}{lrrrr}
    \toprule
        Dataset & Conv. & Chunks & Utterances & Hrs \\
        \midrule
        Callhome/Train & 80 & 164 & 14,996 & 14 \\
        Fisher/Train & 759 & 1,637 & 137,941 & 168 \\
        \midrule
        Callhome/Devtest & 20 & 41 & 3,945 & 4 \\
        Callhome/Evltest & 20 & 23 & 1,826 & 2\\
        \midrule
        Fisher/Dev & 20 & 44 & 3,955 & 5 \\
        Fisher/Dev2 & 20 & 44 & 3,937 & 5\\
        Fisher/Test & 20 & 43 & 3,618 & 4 \\
    \bottomrule
    \end{tabular}
\end{table}

For both the training and the test sets, we present GPT-4 with a conversation and ask it to generate a summary.
Since we know summaries capturing the same content can differ significantly in style, we generate four total summaries for each test conversation by sampling outputs with a temperature of 0.5.
We aim for our evaluation set to be useful even for evaluating models with limited context windows, so we set the maximum number of words in a conversation to 1200 words.
Given this limit, the context, prompt, and a reasonable length summary will all fit within a context window of 2048 tokens. 
If a conversation in either set exceeds the 1200 word limit, we split it into equal-sized chunks and treat each individual chunk as a separate conversation. 
Across both the train and test sets, the summaries range in size from 144 to 443 words, with a median size of 268 words.
We plan to release these reference summaries to the community.

A breakdown of the number of conversations, chunks, utterances, and audio hours for each dataset is shown in Table \ref{tab:baseline-results}.
When building our summarization dataset, each chunk represents one training example or one datapoint for evaluation.
While any of the individual testsets would be sufficient for ASR or MT evaluation, they do not individually provide enough examples for summarization evaluation.
Aggregating all of the test sets gives a total 195 test examples from 100 conversations, a more appropriate number for summarization evaluation.

\subsection{LLM Adaptation}
\label{sec:adapt}

Along with testing off-the-shelf LLMs, we also experiment with supervised fine-tuning for task adaptation in order to understand the potential for improvement and establish strong baselines for future work to compare against. We use LoRA\cite{hu2022lora} finetuning to adapt the models. All fine-tuning experiments are run with 4-bit quantization and fp16 precision. A LoRA adaptor is learned for every linear layer in the model with $r=64$. The training data are GPT-4 reference summaries paired with either the reference English transcripts (Ref) or with the outputs of our Whisper-NLLB speech translation system (MT from ASR). In other words we  we create two training samples for each GPT-4 summary that differ only in the input. We vary which inputs we use during fine-tuning to evaluate the extent to which domain-matched input improves summarization quality. For fine-tuning experiments that make use of both the reference English and MT from ASR inputs we train the model for a single epoch over all training data. When finetuning on only the reference English or MT from ASR transcripts we train for two epochs in order to keep the number of update steps constant across experiments.\footnote{We also ran inference with these models at the one epoch mark and conclude that there is minimal difference when testing at one or two epochs. The ROUGE-L scores in Table \ref{tab:finetune} are the result of two epoch training, while those in the 182 hour condition of Table \ref{tab:finetuneHours} are the result of one epoch of training. The difference in comparable values is less than 0.7 ROUGE.}

\section{Experimental Setup}
\label{sec:setup}

\subsection{ASR Model}
\label{sec:asr}

\begin{table}
    \caption{\label{tab:wer-bleu} {\it WER and BLEU scores for the Whisper ASR model and the NLLB MT model. The last two columns correspond to BLEU scores where the column header refers to the input to the MT system, either output from the Whisper model (ASR-Spanish) or reference transcriptions (Ref-Spanish).}}
    \centering
    \begin{tabular}{lc|cc}
    \toprule
         & & \multicolumn{2}{c}{BLEU} \\
        Test Set & WER & ASR-Spanish & Ref-Spanish \\
        \midrule
        Callhome/Devtest & 29.1 & 21.8 & 30.3 \\
        Callhome/Evltest & 26.6 & 23.0 & 31.2 \\
        Fisher/Dev & 31.9 & 22.6 & 30.3 \\
        Fisher/Dev2 & 32.4 & 23.6 & 32.0 \\
        Fisher/Test & 25.4 & 23.4 & 30.7 \\
    \bottomrule
    \end{tabular}
\end{table}

We use the Whisper-large-v3 model\cite{radford2023robust} for ASR.
While a dataset and language-specific model could likely outperform the Whisper model in the CTS domain, the Whisper model is publicly available and is chosen due to its wide use and reproducibility. 
The WER of the Whisper model on each of the five test sets is shown in the second column of Table \ref{tab:wer-bleu}.
We measure the WER after downcasing the the output and removing punctuation.
This postprocessing is not applied when used in the cascaded pipeline.

\subsection{MT Model}
\label{sec:mt}

We use the NLLB 1.3 Billion parameter dense model \cite{costa2022no} for machine translation.
As with Whisper for ASR, a domain-specific model would likely outperform the NLLB model on CTS, but we use NLLB for better reproducibility.
The BLEU scores for NLLB on the test sets are also shown in Table \ref{tab:wer-bleu}.
We include punctuation when computing the reported BLEU scores, we also tested scoring without punctuation and found it to have negligible impact on the scores so we exclude those results for legibility. 
The third column in Table \ref{tab:wer-bleu} uses the Whisper ASR output as input to NLLB, while the last column uses the reference Spanish transcriptions.
On average the BLEU scores drop by about 8 points when using ASR output as opposed to reference transcriptions.
Note that while some of the test sets contain multiple translations, we only report BLEU scores using a single reference so that the numbers are comparable across test sets.
In the remaining sections we explore the impact of cascaded ASR and MT errors on downstream summarization.

\subsection{LLMs for Summarization}
\label{sec:llm}

As described in Section \ref{sec:summary}, we use GPT-4 to generate the reference summaries.
We then evaluate a range of open-source and API-based models against the GPT-4 generated references.
The API-based models we consider are GPT-3.5 \cite{ouyang2022training} and GPT-4.
The open-source models we consider are the 7 and 13 billion parameter versions of Llama 2 \cite{touvron2023llama2}, the 7 billion parameter Mistral \cite{jiang2023mistral}, and the 45 billion mixture-of-experts model Mixtral-8x7 \cite{jiang2024mixtral}. We focus on these open-source models due to their low compute requirements which make them more amenable to real-world applications. For the same reason all inference is run with 4-bit quantization and fp16 precision. 
All open-source models tested are the officially released chat or instruct tuned versions. It is well-known that the performance of LLMs can vary dramatically depending on the prompt \cite{wei2022chain}.
We follow the guidelines for prompt structure released by the publishers of each of the individual models.
Our exact prompt structure will be released with the reference summaries.
In addition to applying these models off-the-shelf, we also run a set of further supervised fine-tuning experiments with the Mistral 7B model.

\section{Results}
\subsection{Zero-shot}
We evaluate the quality of summaries with ROUGE-L \cite{lin2004automatic}.
We explored a number of other metrics, but most gave a similar ordering in terms of model performance.
While we recognize the pitfalls of focusing on a single metric \cite{fabbri2021summeval}, we only report ROUGE-L due to space concerns.

\begin{table}
    \caption{\label{tab:baseline-results} {\it ROUGE-L scores for summarization using LLMs. For each model, each row represents a different input condition. The first row is the reference translation. The second row is machine translation of the reference transcripts. The final row is the full pipeline, machine translation of the automatic transcripts.}}
    \centering
    \begin{tabular}{lcc|ccc|c}
    \toprule
         & \multicolumn{2}{c|}{Callhome} & \multicolumn{3}{c|}{Fisher} & \\
        Model+Input & Dev & Eval & Dev & Dev2 & Test & All \\
        \midrule
        \multicolumn{7}{l}{\textbf{Llama2-7B}} \\ 
         - Ref. & 23.8 & 23.2 & 25.1 & 23.9 & 24.5 & 24.2 \\
         - MT of Ref. & 23.1 & 22.4 & 23.3 & 22.8 & 24.9 & 23.4 \\
         - MT of ASR & 23.1 & 23.6 & 23.3 & 22.9 & 23.8 & 23.4 \\
         \midrule
        \multicolumn{7}{l}{\textbf{Llama2-13B}} \\
         - Ref. & 23.6 & 23.9 & 25.7 & 25.5 & 26.4 & 25.2\\
         - MT of Ref. & 22.4 & 22.3 & 24.1 & 22.7 & 24.3 & 23.3 \\
         - MT of ASR & 22.8 & 21.9 & 23.9 & 23.6 & 25.3 & 23.7 \\
         \midrule
        \multicolumn{7}{l}{\textbf{Mixtral-8x7B}} \\ 
         - Ref. & 26.7 & 27.0 & 28.4 & 27.7 & 27.6 & 27.5 \\
         - MT of Ref. & 27.1 & 26.9 & 27.8 & 26.9 & 27.2 & 27.2 \\
         - MT of ASR & 26.7 & 25.6 & 27.4 & 26.9 & 27.1 & 26.9 \\
         \midrule
        \multicolumn{7}{l}{\textbf{Mistral-7B}} \\ 
         - Ref. & 24.3 & 22.6 & 26.1 & 25.6 & 24.7 & 24.7\\
         - MT of Ref. & 22.5 & 21.0 & 24.1 & 24.0 & 25.0 & 23.3\\
         - MT of ASR & 22.3 & 21.6 & 24.1 & 23.4 & 23.9 & 23.1\\
         \midrule
        \multicolumn{7}{l}{\textbf{GPT-3.5}} \\
         - Ref. & 24.0 & 23.7 & 28.0 & 27.0 & 27.8 & 26.1\\
         - MT of Ref. & 23.5 & 21.9 & 27.4 & 26.7 & 26.0 & 25.1\\
         - MT of ASR & 20.9 & 20.3 & 25.5 & 25.1 & 25.5 & 23.5\\
         \midrule
        \multicolumn{7}{l}{\textbf{GPT-4}} \\
         - Ref. & --- & --- & --- & --- & --- & ---\\
         - MT of Ref. & 32.0 & 31.7 & 34.1 & 33.0 & 33.7 & 32.9\\
         - MT of ASR & 30.9 & 31.0 & 33.4 & 32.3 & 32.6 & 32.0\\
        \midrule\midrule
        \multicolumn{7}{l}{\textbf{FT Mistral-7B}} \\ 
         - Ref. & 33.1 & 32.6 & 35.1 & 34.5 & 35.2 & 34.3\\
         - MT of Ref. & 32.9 & 32.0 & 34.6 & 33.5 & 34.0 & 33.5\\
         - MT of ASR & 32.2 & 31.0 & 33.3 & 33.0 & 33.4 & 32.7 \\
    \bottomrule
    \end{tabular}
\end{table}

\begin{table}
    \caption{\label{tab:finetune} \it ROUGE-L scores on the reference transcript and MT of ASR variants of the test set from models fine-tuned on reference transcripts, on MT of ASR transcripts, and on both. The \textit{source prompt} version of the fine-tune set includes both reference and MT of ASR transcripts, with a prompt telling the LLM which is which.}
    \centering
    \begin{tabular}{lcc}
    \toprule
        Finetune Data & Ref. Test & MT of ASR Test  \\
        \midrule
        Ref. & 34.7 & 33.1  \\
        MT of ASR & 33.9 & 32.9  \\
        Ref. + MT of ASR & 34.3 & 32.7 \\
         + Source prompt & 34.1 & 33.0  \\
        
    \bottomrule
    \end{tabular}
\end{table}

\begin{table}
    \caption{\label{tab:finetuneHours} \it ROUGE-L scores for the Mistral-7B model on MT of ASR test data after fine-tuning. Each column adds progressively more training data into the finetuning.}
    \centering
    \begin{tabular}{lcccc}
    \toprule
        Train Data & 0 hrs & 32 hrs & 65 hrs & 182 hrs \\
        \midrule
        Ref. Translation & 24.7 & 28.1 & 28.4 & 32.4 \\
        MT of ASR Transcript & 23.9 & 27.2 & 29.1 & 32.3 \\
    \bottomrule
    \end{tabular}
\end{table}

\begin{table*}
    \caption{\label{tab:examples} \it Example summaries highlighting the issues that can be introduced through ASR and MT errors. Both summaries are generated using GPT-4, but the second uses ASR+MT as input. Errors that can be traced to issues with ASR and MT are highlighted in red.}
    \centering
    \begin{tabular}{p{2cm}p{14cm}}
    \toprule
        Summary from Reference Translation & The conversation revolves around family dynamics, work, health, and recent unfortunate events. The initial part of the dialogue discusses the complexities and challenges of working with family members, with both speakers agreeing that it can create discord and unhappiness on both sides. The topic then shifts to personal updates, with one speaker mentioning they and Estela are fine despite gaining weight, which humorously leads to a comment on financial struggles tied to their eating habits. \\
        \midrule
        Summary from ASR+MT & The conversation kicks off with Speaker A expressing frustration that a certain female party is unwilling to work with him, touching on a broader theme of children generally disliking working with their parents. Speaker B offers {\color{red} mundane responses}, possibly indicating detachment or an attempt to change the subject, mentioning the cold weather and previous rough experiences working with families. The dialogue then shifts to personal updates, revealing that Speaker A and his family are dealing with health and weight issues, alongside financial difficulties marked by the mention of someone being {\color{red}"so poor" and possibly "comatose."}...Speaker B, meanwhile, maintains a pattern of responding with {\color{red}"Thank you, I'm all right,"} perhaps to offer polite acknowledgment without engaging deeply with the troubles Speaker A shares. \\
    \bottomrule
    \end{tabular}
\end{table*}

Baseline results are shown in Table \ref{tab:baseline-results}.
Each row represents a different input condition.
While the performance of each model becomes progressively worse as more error is introduced through MT and ASR---as opposed to using reference transcripts and translations---the drop in performance is less than anticipated.
The difference between a summary generated from a reference translation and the cascade of AST and MT is no more than 10\% relative across all models.
We believe there are two possible reasons for this result and they merit future investigation.
Either the errors from transcription and translation do not impact the model's ability to summarize the key information, or the metric is not able to measure the impact.

The performance of the open-source models follows the expected ranking, with the larger MoE model Mixtral outperforming the smaller, dense Mistral and Llama 2 models. Despite the large size difference and the relatively large performance gaps between these models on other benchmarks \cite{jiang2024mixtral}, the actual range of performance across open-source models before fine-tuning is relatively narrow. Mixtral, the best open-source model without fine-tuning, outperforms the worst, Llama2-7B, by only 3.5 BLEU when translating MT of ASR. 

Similarly the models all seem robust to MT and ASR errors even without fine-tuning. The largest gap between ref. and MT of ASR transcript performance is only 1.5 ROUGE. This is in contrast to previous work on other types of cascading systems, like speech MT, where downstream models have been repeatedly shown to be highly sensitive to upstream errors. 

We also compare the open-source models against GPT-3.5. Mixtral outperforms GPT-3.5 across input types and the other models perform competitively. Given the difference in model size and training, the performance of the open-source models is impressive. As a top line for performance, we evaluate GPT-4 generated summaries using both MT of reference transcriptions and MT of ASR.
We do not evaluate GPT-4 on reference translations because it was used to generate the reference summaries.
Even when using the MT input, we expect the results to be biased as it is essentially using the same model to evaluate itself.
GPT-4 obtains a ROUGE-L score almost 50\% higher than some of the competing open-source models.

\subsection{Finetuning}

In addition to testing off-the-shelf API-based and open-source models, we also fine-tune Mistral-7B to provide a much stronger baseline. Table \ref{tab:baseline-results} includes results from a fine-tuned model that is trained with both reference and MT of ASR inputs.
Compared to the unadapted version, fine-tuning improves the performance of the Mistral-7B model by almost 10 ROUGE points.
After fine-tuning, the performance of the model is comparable to, or even outperforms, GPT-4.

In order to determine the value of including both the reference transcript and MT of ASR inputs in fine-tuning, we run a set of experiments in which we vary our fine-tuning dataset. Table \ref{tab:finetune} shows performance, aggregated across datasets, when we fine-tune on only the reference transcripts, only the MT of ASR transcripts, and the combination of both. We also experiment with a fourth condition in which we include both the reference transcripts and MT of ASR transcripts, but use a separate \textit{source prompt} during fine-tuning and inference for the two. The intuition behind this experiment is that there are likely different error distributions between the two types of inputs and it might help the LLM to signal what type of input is being provided. 

We find that ROUGE scores are relatively flat across different fine-tuning sets. Fine-tuning on reference transcripts does result in the model that performs best at summarizing reference transcripts, although the differences are small. The same cannot be said for testing on MT of ASR outputs, where the inclusion of MT of ASR data in the train set does not reliably yield an improvement in ROUGE. This is potentially reflective of the fact that even before fine-tuning models seemed very robust to MT and ASR errors. In Table \ref{tab:finetuneHours} we show how performance varies as we vary the amount of fine-tuning data. We see a roughly linear increase in ROUGE as the amount of data increases.

In Table \ref{tab:examples} we show an example summary using GPT-4 where the input comes either from reference translations or ASR+MT.
The second summary contains several phrases highlighted in red.
We can trace these back to errors in either the ASR or MT.
The statements ```so poor' and `comotose.'" likely arise from a combination of errors and the true statement should be, ``You eat all you earn."
When the summmary mentions Speaker B offering ``mundane responses" and saying, ``Thank you, I'm all right," it is a reflection of hallucinations in both the ASR and the MT.
The Whisper model tends to output ``Gracias" as a filler word and the NLLB model translates it as ``Thank you, I'm all right."
Interestingly, this demonstrates that errors in ASR+MT not only impact the factual information in the summary, but also the implied tone.




\section{Conclusions}

We have established an evaluation framework for CTS summarization.
Using GPT-4, we created reference summaries for a well-known Spanish CTS corpus with existing English translations.
Our experiments establish a baseline for a cascaded approach to summarization using publicly available models.
While GPT-4 outperforms existing open-source models, we are able to match the performance of GPT-4 by fine-tuning the Mistral-7B model.
This demonstrates the efficacy of using large, API-based models like GPT-4 to generate evaluation and adaptation data for cross-lingual speech summarization.

We plan to explore several extensions to this work in the future.
Moving beyond general summarization, we want to explore contextual summarization where the summary can be guided by input from the user to focus on specific information.
This presents further challenges, not just as a task, but also in terms of evaluation.
We also want to incorporate additional information to the summarization system based on alternative hypotheses for both transcription and translation.


\bibliographystyle{IEEEtran}
\bibliography{mybib}

\begin{thebibliography}{10}
\providecommand{\url}[1]{#1}
\csname url@samestyle\endcsname
\providecommand{\newblock}{\relax}
\providecommand{\bibinfo}[2]{#2}
\providecommand{\BIBentrySTDinterwordspacing}{\spaceskip=0pt\relax}
\providecommand{\BIBentryALTinterwordstretchfactor}{4}
\providecommand{\BIBentryALTinterwordspacing}{\spaceskip=\fontdimen2\font plus
\BIBentryALTinterwordstretchfactor\fontdimen3\font minus \fontdimen4\font\relax}
\providecommand{\BIBforeignlanguage}[2]{{%
\expandafter\ifx\csname l@#1\endcsname\relax
\typeout{** WARNING: IEEEtran.bst: No hyphenation pattern has been}%
\typeout{** loaded for the language `#1'. Using the pattern for}%
\typeout{** the default language instead.}%
\else
\language=\csname l@#1\endcsname
\fi
#2}}
\providecommand{\BIBdecl}{\relax}
\BIBdecl

\bibitem{zhang2024benchmarking}
T.~Zhang, F.~Ladhak, E.~Durmus, P.~Liang, K.~McKeown, and T.~B. Hashimoto, ``Benchmarking large language models for news summarization,'' \emph{Transactions of the Association for Computational Linguistics}, vol.~12, pp. 39--57, 2024.

\bibitem{valenza1999summarisation}
R.~Valenza, T.~Robinson, M.~Hickey, and R.~Tucker, ``Summarisation of spoken audio through information extraction,'' in \emph{ESCA Tutorial and Research Workshop (ETRW) on Accessing Information in Spoken Audio}, 1999.

\bibitem{koumpis2000transcription}
K.~Koumpis and S.~Renals, ``Transcription and summarization of voicemail speech,'' in \emph{ICSLP}.\hskip 1em plus 0.5em minus 0.4em\relax International Speech Communication Association, 2000.

\bibitem{hori2002automatic}
C.~Hori, S.~Furui, R.~Malkin, H.~Yu, and A.~Waibel, ``Automatic summarization of english broadcast news speech,'' in \emph{Proceedings of the second international conference on Human Language Technology Research}, 2002, pp. 241--246.

\bibitem{murray05extractive}
G.~Murray, S.~Renals, and J.~Carletta, ``{Extractive summarization of meeting recordings},'' in \emph{Proc. Interspeech 2005}, 2005, pp. 593--596.

\bibitem{sharma2023espnet}
R.~Sharma, W.~Chen, T.~Kano, R.~Sharma, S.~Arora, S.~Watanabe, A.~Ogawa, M.~Delcroix, R.~Singh, and B.~Raj, ``Espnet-summ: Introducing a novel large dataset, toolkit, and a cross-corpora evaluation of speech summarization systems,'' in \emph{2023 IEEE Automatic Speech Recognition and Understanding Workshop (ASRU)}.\hskip 1em plus 0.5em minus 0.4em\relax IEEE, 2023, pp. 1--8.

\bibitem{zou2021topic}
Y.~Zou, L.~Zhao, Y.~Kang, J.~Lin, M.~Peng, Z.~Jiang, C.~Sun, Q.~Zhang, X.~Huang, and X.~Liu, ``Topic-oriented spoken dialogue summarization for customer service with saliency-aware topic modeling,'' in \emph{Proceedings of the AAAI Conference on Artificial Intelligence}, vol.~35, no.~16, 2021, pp. 14\,665--14\,673.

\bibitem{furui2004speech}
S.~Furui, T.~Kikuchi, Y.~Shinnaka, and C.~Hori, ``Speech-to-text and speech-to-speech summarization of spontaneous speech,'' \emph{IEEE Transactions on Speech and Audio Processing}, vol.~12, no.~4, pp. 401--408, 2004.

\bibitem{kano2023summarize}
T.~Kano, A.~Ogawa, M.~Delcroix, K.~Matsuura, T.~Ashihara, W.~Chen, and S.~Watanabe, ``Summarize while translating: Universal model with parallel decoding for summarization and translation,'' in \emph{2023 IEEE Automatic Speech Recognition and Understanding Workshop (ASRU)}.\hskip 1em plus 0.5em minus 0.4em\relax IEEE, 2023, pp. 1--8.

\bibitem{liu2013towards}
F.~Liu and Y.~Liu, ``Towards abstractive speech summarization: Exploring unsupervised and supervised approaches for spoken utterance compression,'' \emph{IEEE Transactions on Audio, Speech, and Language Processing}, vol.~21, no.~7, pp. 1469--1480, 2013.

\bibitem{post2013improved}
M.~Post, G.~Kumar, A.~Lopez, D.~Karakos, C.~Callison-Burch, and S.~Khudanpur, ``Improved speech-to-text translation with the fisher and callhome spanish--english speech translation corpus,'' in \emph{Proc. IWSLT}, 2013.

\bibitem{hu2022lora}
E.~J. Hu, Y.~Shen, P.~Wallis, Z.~Allen-Zhu, Y.~Li, S.~Wang, L.~Wang, and W.~Chen, ``Lora: Low-rank adaptation of large language models,'' in \emph{ICLR}, 2022.

\bibitem{radford2023robust}
A.~Radford, J.~W. Kim, T.~Xu, G.~Brockman, C.~McLeavey, and I.~Sutskever, ``Robust speech recognition via large-scale weak supervision,'' in \emph{International Conference on Machine Learning}.\hskip 1em plus 0.5em minus 0.4em\relax PMLR, 2023, pp. 28\,492--28\,518.

\bibitem{costa2022no}
M.~R. Costa-juss{\`a}, J.~Cross, O.~{\c{C}}elebi, M.~Elbayad, K.~Heafield, K.~Heffernan, E.~Kalbassi, J.~Lam, D.~Licht, J.~Maillard \emph{et~al.}, ``No language left behind: Scaling human-centered machine translation,'' \emph{arXiv preprint arXiv:2207.04672}, 2022.

\bibitem{ouyang2022training}
L.~Ouyang, J.~Wu, X.~Jiang, D.~Almeida, C.~L. Wainwright, P.~Mishkin, C.~Zhang, S.~Agarwal, K.~Slama, A.~Ray \emph{et~al.}, ``Training language models to follow instructions with human feedback,'' 2022.

\bibitem{touvron2023llama2}
H.~Touvron, L.~Martin, K.~Stone, P.~Albert, A.~Almahairi, Y.~Babaei, N.~Bashlykov, S.~Batra, P.~Bhargava, S.~Bhosale \emph{et~al.}, ``Llama 2: Open foundation and fine-tuned chat models,'' \emph{arXiv preprint arXiv:2307.09288}, 2023.

\bibitem{jiang2023mistral}
A.~Q. Jiang, A.~Sablayrolles, A.~Mensch, C.~Bamford, D.~S. Chaplot, D.~d.~l. Casas, F.~Bressand, G.~Lengyel, G.~Lample, L.~Saulnier \emph{et~al.}, ``Mistral 7b,'' \emph{arXiv preprint arXiv:2310.06825}, 2023.

\bibitem{jiang2024mixtral}
A.~Q. Jiang, A.~Sablayrolles, A.~Roux, A.~Mensch, B.~Savary, C.~Bamford, D.~S. Chaplot, D.~d.~l. Casas, E.~B. Hanna, F.~Bressand \emph{et~al.}, ``Mixtral of experts,'' \emph{arXiv preprint arXiv:2401.04088}, 2024.

\bibitem{wei2022chain}
J.~Wei, X.~Wang, D.~Schuurmans, M.~Bosma, F.~Xia, E.~Chi, Q.~V. Le, D.~Zhou \emph{et~al.}, ``Chain-of-thought prompting elicits reasoning in large language models,'' \emph{Advances in neural information processing systems}, vol.~35, pp. 24\,824--24\,837, 2022.

\bibitem{lin2004automatic}
C.-Y. Lin and F.~J. Och, ``Automatic evaluation of machine translation quality using longest common subsequence and skip-bigram statistics,'' in \emph{Proceedings of the 42nd annual meeting of the association for computational linguistics (ACL-04)}, 2004, pp. 605--612.

\bibitem{fabbri2021summeval}
A.~R. Fabbri, W.~Kry{\'s}ci{\'n}ski, B.~McCann, C.~Xiong, R.~Socher, and D.~Radev, ``Summeval: Re-evaluating summarization evaluation,'' \emph{Transactions of the Association for Computational Linguistics}, vol.~9, pp. 391--409, 2021.

\end{thebibliography}

\end{document}